\documentclass[]{IOS-Book-Article}

\usepackage{nesybook}
\usepackage{chapter-sl/chap}

\begin{document}

\newcommand{\paolo}[1]{\textcolor{blue}{\textbf{Paolo:} ``#1''}}
\newcommand{\kareem}[1]{\textcolor{red}{\textbf{Kareem:} ``#1''}}
\newcommand{\guy}[1]{\textcolor{yellow}{\textbf{Guy:} ``#1''}}

\newtheorem{thm}{Theorem}
\newtheorem{lem}{Lemma}
\newtheorem{prop}{Proposition}
\newtheorem{cor}{Corollary}

\chapter{Semantic Loss Functions for Neuro-Symbolic Structured Prediction}
\label{chapter-sl:chap}

\chapterauthor{Kareem Ahmed}{UCLA}
\chapterauthor{Stefano Teso}{UNITN}
\chapterauthor{Paolo Morettin}{KULeuven}
\chapterauthor{Luca Di Liello}{UNITN}
\chapterauthor{Pierfrancesco Ardino}{UNITN}
\chapterauthor{Jacopo Gobbi}{UNITN}
\chapterauthor{Yitao Liang}{Peking University}
\chapterauthor{Eric Wang}{UCLA}
\chapterauthor{Kai-Wei Chang}{UCLA}
\chapterauthor{Andrea Passerini}{UNITN}
\chapterauthor{Guy Van den Broeck}{UCLA}
\allchapterauthors{Kareem Ahmed, Stefano Teso, Paolo Morettin, Luca Di Liello, Pierfrancesco Ardino, Jacopo Gobbi, Yitao Liang, Eric Wang, Kai-Wei Chang, Andrea Passerini, Guy Van den Broeck}

Structured output prediction problems are ubiquitous in machine learning.
The prominent approach leverages neural networks as powerful feature 
extractors, otherwise assuming the independence of the outputs.
These outputs, however, jointly encode an object, e.g. a path in a graph,
and are therefore related through the structure underlying
the output space.
We discuss the \textit{semantic loss}, which injects knowledge about such structure, defined symbolically, into training by minimizing the network's violation of such dependencies, steering the network towards predicting distributions satisfying the  underlying structure.
At the same time, it is agnostic to the arrangement of the symbols, and depends only on
the semantics expressed thereby, while also enabling efficient end-to-end training
and inference.
We also discuss key improvements and applications of the semantic loss.
One limitations of the semantic loss is that it does not exploit the association of every data point with certain features certifying its membership in a target class.
We should therefore prefer \emph{minimum-entropy distributions over valid structures},
which we obtain by additionally minimizing \emph{the neuro-symbolic entropy}.
We empirically demonstrate the benefits of this more refined formulation.
Moreover, the semantic loss is designed to be modular and can be combined
with both discriminative and generative neural models.
This is illustrated by integrating it into generative
adversarial networks, yielding \textit{constrained adversarial networks}, a novel class of deep generative models 
able to efficiently synthesize complex objects obeying
the structure of the underlying domain.

\section{Introduction}
The widespread success of representation learning raises the question of which AI
tasks are amenable to deep learning, which tasks require classical model-based 
symbolic reasoning, and whether we can benefit from a tighter integration of both 
approaches.
Here we consider learning in domains where we have symbolic knowledge connecting 
the different outputs of a neural network. This knowledge takes the form of a 
constraint (or sentence) in Boolean logic. It can be as simple as an exactly-one 
constraint for one-hot output encodings, or as complex as a structured output 
prediction constraint for intricate combinatorial objects such as rankings, 
subgraphs, or paths. Our goal is to augment neural networks with the ability to 
learn how to make predictions subject to these constraints, and use the symbolic 
knowledge to improve the learning performance.

Most neuro-symbolic approaches aim to simulate or learn symbolic reasoning in an
end-to-end deep neural network, or capture symbolic knowledge in a vector-space 
embedding.
This choice is partly motivated by the need for smooth \emph{differentiable} 
models; adding symbolic reasoning code (e.g., SAT solvers) to a deep learning
pipeline destroys this property.
Unfortunately, while making reasoning differentiable, the precise logical 
meaning of the knowledge is often lost. In this chapter, we take a distinctly
unique approach, and tackle the problem of differentiable but sound logical
reasoning from first principles. Starting from a set of intuitive axioms, 
we derive the differentiable \emph{semantic loss} which captures how well
the outputs of a neural network match a given constraint. This function 
captures the \emph{meaning} of the constraint, and is independent of its
\emph{syntax}.

Semantic loss \cite{Xu18} minimizes the probability allocated by the network to invalid outputs,
but does not further restrict the shape of the output distribution.
A given data point is associated with certain features, certifying it's membership in 
a specific underlying class.
Intuitively, a classifier guessing uniformly at random has \emph{maximum entropy} and 
has not learned features that are informative of the underlying class.
Therefore, we should prefer learning \emph{minimum-entropy} distributions that satisfy the 
constraints. 
Naively, we might consider simply optimizing both losses simultaneously. 
Computed in that manner, entropy regularization does not account for the 
structure of the output space and is therefore likely to push the network
towards invalid outputs.
\emph{Neuro-symbolic entropy regularization} \cite{Ahmed2022UAI} restricts the entropy loss to the network's
distribution over the valid outputs, as opposed to the entire predictive distribution.
That is, we require that the network's output distribution be \emph{maximally informative} 
of the target \emph{subject to the constraint}.
Computing the entropy of a distribution subject to a constraint is computationally hard.
We provide an algorithm leveraging structural properties of logical circuits to efficiently
compute it.

Empirically, we evaluate \emph{semantic loss} and \emph{neuro-symbolic entropy} on four 
structured output prediction tasks, in both semi- and fully-supervised settings, leading
to models whose predictions are more accurate, and more likely to satisfy the constraint.

Many key applications, however, require \emph{generating} objects that satisfy hard 
structural constraints, like drug molecules, which must be chemically valid, and game
levels, which must be playable.
Despite their impressive success~\cite{karras2018progressive,zhang2017stackgan,zhu2017unpaired}, Generative Adversarial Networks (GANs) ~\cite{goodfellow2014generative}
struggle in these applications, the data often being insufficient or insufficiently well distributed to capture the structural
constraints -- especially if noisy -- and convey them to the model.
\textit{Constrained Adversarial Networks} (CANs) extend GANs to generating valid structures with high probability.
Given a set of arbitrary symbolic constraints, CANs achieve this by penalizing
the generator for allocating mass to invalid objects during training, implemented using 
semantic loss.
CANs are able to handle complex constraints, like reachability on graphs, by first embedding
the candidate configurations in a space in which the constraints can be  encoded compactly,
and then applying the SL to the embeddings.
Since the constraints are embedded directly into the generator, high-quality structures can be
sampled efficiently (in time practically independentof the complexity of the constraints) with
a simple forward pass on the generator, as in regular GANs.
No costly sampling or optimization steps are needed.
We also show how to equip CANs with the ability to switch constraints on and off dynamically
during inference, at no runtime cost.

\subsection{Notation}\label{sec:background}
We write uppercase letters ($X$, $Y$) for Boolean variables and lowercase letters ($x$, $y$) for their instantiation ($Y=0$ or $Y=1$).
Sets of variables are written in bold uppercase ($\Xs$, $\Ys$), and their joint instantiation in bold lowercase ($\xs$, $\ys$).
A literal is a variable ($Y$) or its negation ($\neg Y$).
A logical sentence ($\alpha$ or $\beta$) is constructed from variables and logical connectives ($\land$, $\lor$, etc.), and is also called a (logical) formula or constraint.
A state or world $\ys$ is an instantiation to all variables $\Ys$.
A state $\ys$ satisfies a sentence $\alpha$, denoted $\ys \models \alpha$, if the sentence evaluates to true in that world. A state $\ys$ that satisfies a sentence $\alpha$ is also said to be a model of $\alpha$.
We denote by $m(\alpha)$ the set of all models of $\alpha$.
The notation for states $\ys$ refers to an assignment, the logical sentence enforcing the assignment, or the binary output vector capturing the assignment, all equivalent notions.
A sentence $\alpha$ entails another sentence $\beta$, denoted $\alpha \models \beta$, if all worlds that satisfy $\alpha$ also satisfy $\beta$.

\section{Semantic Loss}
\label{sec:semantic_loss}
Let $\alpha$ be a logical sentence defined over Boolean variables $\Ys = \{Y_1,\dots,Y_n\}$.
Let $\pv$ be a vector of probabilities for the same variables $\Ys$, where $\pv_i$ denotes the predicted probability of variable $Y_i$ and corresponds to a single output of the neural network.
The neural network's outputs induce a probability distribution $\prob(\cdot)$ over all possible states $\ys$ of $\Ys$:
\begin{equation}\label{eqn:pr_struct}
     \prob(\y) = \prod_{i: \y \models \Y_i} \pv_i \prod_{i: \y \models \lnot \Y_i} (1 - \pv_i).
\end{equation}

The \textit{semantic loss} is a function of the logical constraint $\alpha$ and a probability vector $\pv$. 
It quantifies how close the neural network comes to satisfying the constraint by computing the probability of the constraint under the distribution $\prob(\cdot)$ induced by $\pv$.
It does so by reducing the problem of probability computation to weighted model counting (WMC): summing up the models of $\alpha$, each weighted by its likelihood under $\prob(\cdot)$.
It, therefore, maximizes the probability mass allocated by the network to the models of $\alpha$:
\begin{equation}
\label{eq:sloss}
\mathbb{E}_{\y \sim \prob}\left[ \Ind{\y \models \alpha} \right] 
= \sum_{\y \models \alpha} \prob(\y).
\end{equation}
Here, $\Ind{cond}$ is the indicator function that evaluates to $1$ whenever $cond$ holds and to $0$ otherwise.
Taking the negative logarithm recovers semantic loss $\SL{\alpha}{\prob}$.
Intuitively, the semantic loss is proportional to (the negative logarithm of) the
probability of generating a state that satisfies the constraint, when sampling
values according to $\pv$. Hence, it is the self-information (or ``surprise'')
of obtaining an assignment that satisfies the constraint~\cite{jones1979elementary}.
The quantity in Equation \ref{eq:sloss} is generally \#P-hard to compute
~\cite{Valiant1979a, Valiant1979b}.
Next, we will show that, through compiling the logical formula into tractable circuits
satisfying certain structural properties, we can compute the above quantity in time 
that is linear in the size of the circuit.

\subsection{Tractable Computation through Knowledge Compilation}

We resort to knowledge compilation techniques -- a class of methods that transform, or \emph{compile}, a logical theory into a target form with certain properties that allow certain probabilistic queries to be answered efficiently.

\paragraph{Logical Circuits} 
More formally, a \emph{logical circuit} is a directed, acyclic computational graph representing a logical formula.
Each node $n$ in the DAG encodes a logical sub-formula, denoted $[n]$.
Each inner node in the graph is either an AND or an OR gate, and each leaf node encodes a Boolean literal ($Y$ or $\lnot Y$). 
We denote by $\ch(n)$ the set of $n$'s children, that is, the operands of its logical gate.

\paragraph{Structural Properties} Circuits enable the tractable computation of certain classes of queries over encoded functions granted that a set of structural properties are enforced. %

A circuit is \emph{decomposable} if the inputs of every AND gate depend on disjoint sets of variables i.e. for $\alpha = \beta \land \gamma$, $\vars(\beta) \cap \vars(\gamma) = \varnothing$.
Intuitively, decomposable AND nodes encode local factorizations of the function. For the sake of simplicity, we assume that decomposable AND gates always have two inputs, a condition that can be enforced on any circuit in exchange for a polynomial increase in its size~\cite{vergari2015simplifying,peharz2020einsum}.

A second useful property is \emph{smoothness}.
A circuit is \emph{smooth} if the children of every OR gate depend on the same set of variables, i.e. for $\alpha = \bigvee_i \beta_i$, we have that $\vars(\beta_i) = \vars(\beta_j)\ \forall i,j$. Decomposability and smoothness are a sufficient and necessary condition for tractable integration over arbitrary sets of variables in a single pass, as they allow larger integrals to decompose into smaller ones~\cite{choi2020pc}.

Lastly, a circuit is said to be  \emph{deterministic} if, for any input, at most one child of every OR node has a non-zero output, i.e. for $\alpha = \bigvee_i \beta_i$, we have that $ \beta_i \land \beta_j = \bot$ for all $i \neq j$. Figure \ref{fig:example} shows an example of smooth, decomposable and deterministic circuit.
Taken together, decomposability and determinism are a necessary and sufficient condition, for computing Equation \ref{eq:sloss}, as well as its gradients with respect to the network's weights, in time linear in the size of the circuit~\cite{darwiche02,VergariNeurIPS21}.
This does not in general escape the complexity of the computation: worst case, the compiled circuit can be exponential in the size of the constraint.
In practice, constraints exhibit enough structure, like repeated subproblems, to make compilation feasible.

So far we have covered the basics needed to understand semantic loss functions and their pratical implementation.
We proceed by discussing an improved semantic loss that takes into consideration the entropy of the learned distribution in Section~\ref{sec:nesyent}, and then discuss applications of the semantic loss to the context of deep generative modeling of structured outputs in Section~\ref{sec:can}.

\section{Neuro-Symbolic Entropy Regularization}
\label{sec:nesyent}

\subsection{Motivation and Definition}

\begin{figure}[t]
\begin{subfigure}[b]{0.23\textwidth}
\centering
\begin{tikzpicture}
\begin{axis}[
    no markers, domain=0:10, samples=100,
    axis lines*=left, xlabel=$\y$, ylabel=$p(\y|x)$,
    every axis y label/.style={at=(current axis.above origin),anchor=south},
    every axis x label/.style={at=(current axis.right of origin),anchor=west},
    height=3cm,
    xtick=\empty, ytick=\empty,
    enlargelimits=false, clip=false, axis on top,
    grid = major,
    ]
    \addplot [fill=cyan!20, draw=none, domain=1.5:8.5] {gaussianmixture2(0.5,2,7,1,0.5)} \closedcycle;
    \addplot [very thick,cyan!50!black] {gaussianmixture2(0.5,2,7,1,0.5)};
    \draw [yshift=-0.3cm, latex-latex, |-|](axis cs:1.5,0) -- node [fill=white] {$m(\alpha)$} (axis cs:8.5,0);
\end{axis}
\end{tikzpicture}
\caption{Network uncertain over both valid as well as invalid predictions}
\end{subfigure}
\hfill
\begin{subfigure}[b]{0.23\textwidth}
\centering
\begin{tikzpicture}
\begin{axis}[
    no markers, domain=0:10, samples=500,
    axis lines*=left, xlabel=$\y$, ylabel=$p(\y|x)$,
    every axis y label/.style={at=(current axis.above origin),anchor=south},
    every axis x label/.style={at=(current axis.right of origin),anchor=west},
    height=3cm,
    xtick=\empty, ytick=\empty,
    enlargelimits=false, clip=false, axis on top,
    grid = major,
    ]
    \addplot [fill=cyan!20, draw=none, domain=1.5:8.5] {gaussianmixture2(0.6,0.2,4,2.5,0.6)} \closedcycle;
    \addplot [very thick,cyan!50!black] {gaussianmixture2(0.6,0.2,4,2.5,0.6)};
    \draw [yshift=-0.3cm, latex-latex, |-|](axis cs:1.5,0) -- node [fill=white] {$m(\alpha)$} (axis cs:8.5,0);
\end{axis}
\end{tikzpicture}
\caption{Network that is allocating most mass to one invalid prediction}
\end{subfigure}
\hfill
\begin{subfigure}[b]{0.23\textwidth}
\centering
\begin{tikzpicture}
\begin{axis}[
    no markers, domain=0:10, samples=100,
    axis lines*=left, xlabel=$\y$, ylabel=$p(\y|x)$,
    every axis y label/.style={at=(current axis.above origin),anchor=south},
    every axis x label/.style={at=(current axis.right of origin),anchor=west},
    height=3cm,
    xtick=\empty, ytick=\empty,
    enlargelimits=false, clip=false, axis on top,
    grid = major,
    ]
    \addplot [fill=cyan!20, draw=none, domain=1.5:8.5] {gaussianmixture2(3,0.5,6,1,0.5)} \closedcycle;
    \addplot [very thick,cyan!50!black] {gaussianmixture2(3,0.5,6,1,0.5)};
    \draw [yshift=-0.3cm, latex-latex, |-|](axis cs:1.5,0) -- node [fill=white] {$m(\alpha)$} (axis cs:8.5,0);
\end{axis}
\end{tikzpicture}
\caption{Network that is allocating most mass to valid predictions}
\end{subfigure}
\hfill
\begin{subfigure}[b]{0.23\textwidth}
\centering
\begin{tikzpicture}
\begin{axis}[
    no markers, domain=0:10, samples=500,
    axis lines*=left, xlabel=$\y$, ylabel=$p(\y|x)$,
    every axis y label/.style={at=(current axis.above origin),anchor=south},
    every axis x label/.style={at=(current axis.right of origin),anchor=west},
    height=3cm,
    xtick=\empty, ytick=\empty,
    enlargelimits=false, clip=false, axis on top,
    grid = major,
    ]
    \addplot [fill=cyan!20, draw=none, domain=1.5:8.5] {gaussianmixture2(3,0.2,4,2.5,0.6)} \closedcycle;
    \addplot [very thick,cyan!50!black] {gaussianmixture2(3,0.2,4,2.5,0.6)};
    \draw [yshift=-0.3cm, latex-latex, |-|](axis cs:1.5,0) -- node [fill=white] {$m(\alpha)$} (axis cs:8.5,0);
\end{axis}
\end{tikzpicture}
\caption{Network that is allocating most mass to one valid prediction}
\end{subfigure}
\caption{%
A network's predictive distribution can be uncertain or certain ($\leftrightarrow$), and it can allow or disallow invalid predictions under the constraint $\alpha$ ($\updownarrow$).
Entropy regularization steers the network towards confident, possibly invalid predictions (b). 
Neuro-symbolic learning steers the network towards valid predictions without necessarily being confident (c).
Neuro-symbolic entropy-regularization guides the network to valid and confident predictions~(d).}
\label{fig:entsl}
\end{figure}
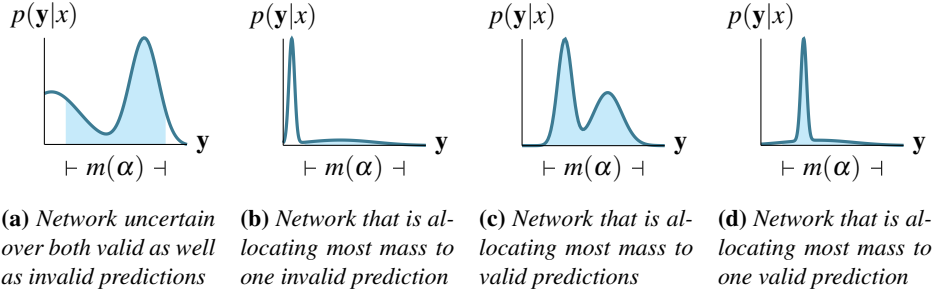

Consider the plots in Figure \ref{fig:entsl}.
For any given data point $x$, the neural network can be fairly uncertain regarding the target class, accommodating for both valid and invalid structured predictions under its predicted distribution.

A common underlying assumption in many machine learning methods is that data belonging to the same class tend to form discrete clusters\;\cite{ssl} -- an assumption deemed justified on the sheer basis of the existence of classes.
Consequently, a classifier is expected to favor decision boundaries lying in regions of low data density, separating the clusters.
Entropy-regularization \cite{grandvalet2005} directly implements the above assumption, requiring that the classifier output confident -- low-entropy -- predictive distributions, pushing the decision boundary away from unlabeled points, thereby supplementing scarce labeled data with abundant unlabeled data.
Seen through that lens, minimizing the entropy of the predictive distribution can be regarded as minimizing a measure of class overlap as a function of the features learned by the network.

Entropy regularization, however, remains agnostic to the underlying domain, failing to exploit situations where we have knowledge characterizing valid predictions in the domain.  
Therefore, it can often be detrimental to a model's performance, causing it to grow confident in invalid predictions.

Conversely, neuro-symbolic approaches steer the network towards distributions disallowing invalid predictions, by maximizing the constraint probability, but do little to ensure the network learn features conducive to classification.

Clearly then, there is a benefit to combining the merits of both approaches. We restrict the entropy computation to the distribution over models of the logical formula, ensuring the network only grow confident in valid predictions. Complemented with maximizing the constraint probability, the network learns to allocate all of its mass to models of the constraint, while being maximally informative of the target.

\paragraph{Defining the Loss}
More precisely, let $\Y$ be a random variable distributed according to Equation \ref{eqn:pr_struct}: $\Y\sim \prob$. We are interested in minimizing the entropy of $\Y$ conditioned on the constraint $\alpha$
\begin{equation}
    \begin{aligned}\label{eqn:nsentropy}
    H(\Y | \alpha)  = - \sum_{\y \models \alpha} \prob(\y | \alpha) \log \prob(\y | \alpha)
                    = - \mathbb{E}_{\Y | \alpha} \left[ \log \prob(\Y | \alpha) \right].
    \end{aligned}
\end{equation}

\subsection{Computing the Loss}\label{sec:compute_nser}
The above loss is, in general, hard to compute. 
To see this, consider the uniform distribution over models of a constraint $\alpha$.
That is, let $\prob(\y|\alpha) = \frac{1}{|m(\alpha)|}$ for all $\y \models \alpha$.
Then, $H(\Y | \alpha) = -\sum_{\y \models \alpha} \frac{1}{|m(\alpha)|} \log \frac{1}{|m(\alpha)|} = \log |m(\alpha)|$.
This tells us how many models of $\alpha$ there are, which is a well-known \#P-hard problem, as we've established in Section \ref{sec:semantic_loss}.

\begin{algorithm}[t]
  \caption{\textsc{Ent}($\alpha, \prob, \cache$)}
  \label{alg:Shannon-Entropy}
  {\bfseries Input:} a smooth, deterministic and decomposable logical circuit $\alpha$, a fully-factorized probability distribution $\prob(\cdot)$ over states of $\alpha$, and a cache $\cache$ for memoization\\
  {\bfseries Output:} $H(\Y | \alpha)$, where $\Y \sim \prob(\cdot)$
   
  \begin{algorithmic}[1]
  \LineIf{$\alpha \in \mathsf{c}$}{\textbf{return} $\cache(\alpha)$}
  \If{$\alpha$ is a literal}
    \State $e \leftarrow 0$
  \ElsIf{$\alpha$ is an AND gate}
  \State $e \leftarrow \entropy(\beta, \prob, \cache) + \entropy(\gamma, \prob, \cache)$
  \ElsIf{$\alpha$ is an OR gate}
  \State $e\leftarrow\sum_{i=1}^{|\ch(\alpha)|} \prob(\beta_i) \,  \entropy(\beta_i, \prob, \cache) - \prob(\beta_i) \log \prob(\beta_i)$
  \EndIf
  \State $\mathsf{c}(\alpha)\leftarrow e$
  \State \textbf{return} $e$
\end{algorithmic}
\end{algorithm}

Let $\alpha$ be a \emph{smooth}, \emph{deterministic} and \emph{decomposable} logical circuit encoding our constraint, defined over Boolean variables $\Ys = \{Y_1,\dots,Y_n\}$. 
We now show that we can compute the constrained entropy in Equation~\ref{eqn:nsentropy} in time linear in the size of $\alpha$.
The key insight is that, using circuits, we are able to efficiently decompose an expectation with respect to a fully-factorized distribution by alternately splitting the query variables and the support of the distribution until we reach the leaves of the circuit, which are simple literals. In what follows, in a slight abuse of notation for brevity, all unconditional probabilities are implicitly conditioned on constraint $\alpha$; that is we redefine $\prob(\cdot)$  as  $\prob(\cdot | \alpha)$.

\paragraph{Base Case: $\alpha$ is a literal}
When $\alpha$ is a literal, $\alpha = Y_i$ or $\alpha = \lnot Y_i$, we have that
\begin{align*}
    \prob(y_i|\alpha) = \Ind{y_i \models [\alpha]}, \text{~~and~~}
    H(y_i | \alpha) = - \prob(y_i|\alpha) \log \prob(y_i|\alpha) = 0.
\end{align*}
Intuitively, a literal has no uncertainty associated with it.
\paragraph{Recursive Case: $\alpha$ is a conjunction}
When $\alpha$ is a conjunction, decomposability enables us to write
\begin{equation*}
    \prob(\y|\alpha) = \prob(\y_1|\beta) \prob(\y_2|\gamma), \text{~where~}  \vars(\beta) \cap \vars(\gamma) = \varnothing
\end{equation*}
as it decomposes 
    $\alpha$ into two independent constraints $\beta$ and~$\gamma$,
    and $\y$ into two independent assignments $\y_1$ and~$\y_2$.
The neuro-symbolic entropy $- \mathbb{E}_{\Y | \alpha} \left[ \log \prob(\Y | \alpha) \right]$ is then
\begin{align*}
    &- \mathbb{E}_{\{\Y_1,\Y_2\} | \alpha} \Big[ \log \prob(\Y_1|\beta) + \log \prob(\Y_2|\gamma)\Big]
    =- \Big[\mathbb{E}_{\Y_1 | \beta} \big[ \log \prob(\Y_1|\beta) \big] + \mathbb{E}_{\Y_2 | \gamma} \big [\log \prob(\Y_2|\gamma)\big]\Big].
\end{align*}
That is, the entropy given a decomposable conjunction $\alpha$ is the sum of entropies given the conjuncts of~$\alpha$.
\paragraph{Recursive Case: $\alpha$ is a disjunction}
When $\alpha$ is a smooth and deterministic disjunction, 
we have that $\alpha = \bigvee_i \beta_i$, where the $\beta_i$s are mutually exclusive, and therefore partition $\alpha$. Consequently, we have that
\begin{equation*}
    \prob(\y|\alpha) = \sum_i \prob(\beta_i) \cdot \prob(\y|\beta_i).
\end{equation*}
The neuro-symbolic entropy decomposes as well:
\begingroup
\allowdisplaybreaks
\begin{align*}
    &- \mathbb{E}_{\Y | \alpha} \left[ \log \prob(\Y | \alpha) \right] 
    = -\sum_{\y \models \alpha}\prob(\y|\alpha)\log \prob(\y|\alpha)\\
    &= -\sum_{\y \models \alpha} \sum_i \prob(\beta_i)\prob(\y|\beta_i) \log \Big[\sum_j \prob(\beta_j)\prob(\y|\beta_j)\Big]\\
    \begin{split}
      &= -\sum_{\y \models \alpha} \sum_i \prob(\beta_i)\prob(\y|\beta_i)\Ind{\y \models \beta_i} 
      \log \Big[\sum_j \prob(\beta_j)\prob(\y|\beta_j)\Ind{\y \models \beta_j}\Big],
    \end{split}
    \intertext{where by determinism, we have that, for any $\y$ such that $\y \models \alpha$, $\y \models \beta_i \implies \y \not \models \beta_j$ for all $i \neq j$. In other words, any state that satisfies the constraint $\alpha$ satisfies one and only one of its terms, and therefore, the above expression equals}
    &-\sum_{\y \models \alpha} \sum_i \prob(\beta_i) \prob(\y|\beta_i)\log \Big[\prob(\beta_i)\prob(\y|\beta_i)\Big]\Ind{\y \models \beta_i}\\
    &= -\sum_i \sum_{\y \models \beta_i} \prob(\beta_i) \prob(\y|\beta_i)\log \Big[\prob(\beta_i)\prob(\y|\beta_i)\Big].\\
    \intertext{Further simplifying the expression, expanding the logarithm, and using the fact that probability sums to 1 yields}
    \begin{split}
        &=-\sum_i \prob(\beta_i) \log \prob(\beta_i) \sum_{\y \models \beta_i} \prob(\y|\beta_i) + \prob(\beta_i) \sum_{\y \models \beta_i} \prob(\y|\beta_i)\log \prob(\y|\beta_i)
    \end{split}\\
    &=-\sum_i \prob(\beta_i) \log \prob(\beta_i) + \prob(\beta_i) \mathbb{E}_{\Y | \beta_i} \Big[ \log \prob(\Y|\beta_i) \Big].
\end{align*}
\endgroup
That is, the entropy of the random variable $\Y$ conditioned on a disjunction $\alpha$ is the sum of the entropy of the distribution induced on the children of $\alpha$, and the average entropy of its children. The full algorithm is illustrated in Algorithm \ref{alg:Shannon-Entropy}.

\begin{figure}[t!]
 \begin{subfigure}[b]{0.33\textwidth}
\scalebox{0.7}{
\begin{tikzpicture}[circuit logic US]
\node (output) at (4.25, 8) {};
\node (or1) [or gate, inputs=nn, rotate=90, scale=0.9] at (4.25,7) {};
\draw (or1) -- (output) node[pos=0.2, above right, color=YellowOrange] {$0.88$};

\node (and1) [and gate, inputs=nn, rotate=90, scale=0.9] at (3,5.8) {};
\node (and2) [and gate, inputs=nn, rotate=90, scale=0.9] at (5.7,5.8) {};

\node (c) at (2.2,4.6) {$C$};
\node (cval) at (2.2,3.8) {$\color{red}{0.2}$};
\draw[->] (cval) edge (c);
\node (nc) at (7,4.6) {$\neg C$};
\node (ncval) at (7,3.8) {$\color{red}0.8$};
\draw[->] (ncval) edge (nc);
\node (or2) [or gate, inputs=nnn, rotate=90, scale=0.9] at (3.1,4.6) {};
\node (or3) [or gate, inputs=nn, rotate=90, scale=0.9] at (5.6,4.6) {};

\node (and3) [and gate, inputs=nn, rotate=90, scale=0.9] at (3.1,3.4) {};
\node (and4) [and gate, inputs=nn, rotate=90, scale=0.9] at (5.1,3.4) {};
\node (and5) [and gate, inputs=nn, rotate=90, scale=0.9] at (6.1,3.4) {};

\node (a) at (4.6,2.2) {$A$};
\node (aval) at (4.6,1.4) {$\color{blue}0.3$};
\draw[->] (aval) edge (a);
\node (nb) at (5.2,2.2) {$\neg B$};
\node (nbval) at (5.2,1.4) {$\color{ForestGreen}0.5$};
\draw[->] (nbval) edge (nb);
\node (na) at (6,2.2) {$\neg A$};
\node (naval) at (6,1.4) {$\color{blue}0.7$};
\draw[->] (naval) edge (na);

\node (or4) [or gate, inputs=nn, rotate=90, scale=0.9] at (2.6,2.2) {};
\node (or5) [or gate, inputs=nn, rotate=90, scale=0.9] at (3.6,2.2) {};
\node (a1) at (2.1,1) {$A$};
\node (a1val) at (2.1,0.2) {$\color{blue}0.3$};
\draw[->] (a1val) edge (a1);
\node (na1) at (2.7,1) {$\neg A$};
\node (na1val) at (2.7,0.2) {$\color{blue}0.7$};
\draw[->] (na1val) edge (na1);
\node (b) at (3.5,1) {$B$};
\node (bval) at (3.5,0.2) {$\color{ForestGreen}0.5$};
\draw[->] (bval) edge (b);
\node (nb1) at (4.1,1) {$\neg B$};
\node (nb1val) at (4.1,0.2) {$\color{ForestGreen}0.5$};
\draw[->] (nb1val) edge (nb1);

\draw (or1.input 1) -- ++ (down: 0.25) -| (and1) node[pos=0.45, above right, color=YellowOrange] {$0.2$};
\draw (or1.input 2) -- ++ (down: 0.25) -| (and2) node[pos=0.1, above right, color=YellowOrange] {$0.68$};

\draw (and1.input 1) -- ++ (down: 0.25) -| (c);
\draw (and1.input 2) -- (or2) node[pos=0.45, right, color=YellowOrange] {$1$};

\draw (and2.input 2) -- ++ (down: 0.25) -| (nc);
\draw (and2.input 1) -- (or3) node[pos=0.45, left, color=YellowOrange] {$0.85$};

\draw (or2.input 2) -- (and3) node[pos=0.45, right, color=YellowOrange] {$1$};

\draw (or3.input 1) -- ++ (down: 0.25) -| (and4) node[pos=0.45, left, color=YellowOrange] {$0.15$};
\draw (or3.input 2) -- ++ (down: 0.25) -| (and5) node[pos=0.45, right, color=YellowOrange] {$0.7$};

\draw (and3.input 1) -- ++ (down: 0.25) -| (or4) node[pos=0.45, above, color=YellowOrange] {1};
\draw (and3.input 2) -- ++ (down: 0.25) -| (or5) node[pos=0.45, above, color=YellowOrange] {1};

\draw (and4.input 1) -- ++ (down:0.4) -| (a);
\draw (and4.input 2) edge (nb);

\draw (and5.input 1) edge (na);
\draw (and5.input 2) -- ++ (down:0.25) -| (or5);

\draw (or4.input 1) -- ++ (down:0.25) -| (a1);
\draw (or4.input 2) edge (na1);

\draw (or5.input 1) edge (b);
\draw (or5.input 2) -- ++ (down:0.25) -| (nb1);
\end{tikzpicture}
}
\end{subfigure}
\begin{subfigure}[b]{0.20\textwidth}
\scalebox{.5}{
\begin{tikzpicture}[baseline=-220   pt]
\node[circle,draw,minimum size=0.8cm] (a1) at (0,0.5) {};
\node[circle,draw,minimum size=0.8cm] (a2) at (0,1.5) {};
\node[circle,draw,minimum size=0.8cm] (a3) at (0,2.5) {};

\node[circle,draw,minimum size=0.8cm] (b1) at (1.5,0) {};
\node[circle,draw,minimum size=0.8cm] (b2) at (1.5,1) {};
\node[circle,draw,minimum size=0.8cm] (b3) at (1.5,2) {};
\node[circle,draw,minimum size=0.8cm] (b4) at (1.5,3) {};

\foreach \i in {1,...,3} {
    \foreach \j in {1,...,4} {
        \draw (a\i) -- (b\j);
    }
}

\node[circle,draw,minimum size=0.8cm] (c1) at (3.0,0) {};
\node[circle,draw,minimum size=0.8cm] (c2) at (3.0,1) {};
\node[circle,draw,minimum size=0.8cm] (c3) at (3.0,2) {};
\node[circle,draw,minimum size=0.8cm] (c4) at (3.0,3) {};

\foreach \i in {1,...,4} {
    \foreach \j in {1,...,4} {
        \draw (b\i) -- (c\j);
    }
}

\node[circle,draw,minimum size=0.8cm] (d1) at (4.5,0.5) {$\color{blue}0.3$};
\node[circle,draw,minimum size=0.8cm] (d2) at (4.5,1.5) {$\color{ForestGreen}0.5$};
\node[circle,draw,minimum size=0.8cm] (d3) at (4.5,2.5) {$\color{red}0.2$};

\foreach \i in {1,...,4} {
    \foreach \j in {1,...,3} {
        \draw (c\i) -- (d\j);
    }
}
\end{tikzpicture}
}
\end{subfigure}
\begin{subfigure}[b]{0.33\textwidth}
\scalebox{0.7}{
\begin{tikzpicture}[circuit logic US]
\node (output) at (4.25, 8) {};
\node (or1) [or gate, inputs=nn, rotate=90, scale=0.9] at (4.25,7) {};
\node (or1val) [color=DarkOrchid] at ($(or1) + (0.7,0.1)$) {$\color{DarkOrchid}1.64$};
\draw (or1) -- (output);

\node (and1) [and gate, inputs=nn, rotate=90, scale=0.9] at (3,5.8) {};
\node (and1val) [color=DarkOrchid] at ($(and1) + (0.7,0.1)$) {$1.30$};
\node (and2) [and gate, inputs=nn, rotate=90, scale=0.9] at (5.7,5.8) {};
\node (and2val) [color=DarkOrchid] at ($(and2) + (0.7,0.1)$) {$1.04$};

\node (c) at (2.2,4.6) {$C$};
\node (cval) [color=DarkOrchid] at ($(c) - (0,0.4)$) {$0$};
\node (nc) at (7,4.6) {$\neg C$};
\node (ncval) [color=DarkOrchid] at ($(nc) - (0,0.4)$) {$0$};
\node (or2) [or gate, inputs=nnn, rotate=90, scale=0.9] at (3.1,4.6) {};
\node (or2val) [color=DarkOrchid] at ($(or2) + (0.7,0.1)$) {$1.30$};
\node (or3) [or gate, inputs=nn, rotate=90, scale=0.9] at (5.6,4.6) {};
\node (or3val) [color=DarkOrchid] at ($(or3) + (-0.7,0.1)$) {$1.04$};

\node (and3) [and gate, inputs=nn, rotate=90, scale=0.9] at (3.1,3.4) {};
\node (and3val) [color=DarkOrchid] at ($(and3) + (-0.7,0.1)$) {$1.30$};
\node (and4) [and gate, inputs=nn, rotate=90, scale=0.9] at (5.1,3.4) {};
\node (and4val) [color=DarkOrchid] at ($(and4) + (-0.6,0.1)$) {$0$};
\node (and5) [and gate, inputs=nn, rotate=90, scale=0.9] at (6.1,3.4) {};
\node (and5val) [color=DarkOrchid] at ($(and5) + (0.7,0.1)$) {$0.69$};

\node (a) at (5,1.7) {$A$};
\node (aval) [color=DarkOrchid] at ($(a) - (0,0.4)$) {$0$};
\node (nb) at (5.8,1.7) {$\neg B$};
\node (nbval) [color=DarkOrchid] at ($(nb) - (0,0.4)$) {$0$};
\node (na) at (6.6,1.7) {$\neg A$};
\node (naval) [color=DarkOrchid] at ($(na) - (0,0.4)$) {$0$};

\node (or4) [or gate, inputs=nn, rotate=90, scale=0.9] at (2.6,2.2) {};
\node (or4val) [color=DarkOrchid] at ($(or4) + (-0.7,0.1)$) {$0.61$};
\node (or5) [or gate, inputs=nn, rotate=90, scale=0.9] at (3.6,2.2) {};
\node (or5val) [color=DarkOrchid] at ($(or5) + (0.8,0.1)$) {$0.69$};

\node (a1) at (2.0,1) {$A$};
\node (a1val) [color=DarkOrchid] at ($(a1) - (0,0.4)$) {$0$};
\node (na1) at (2.7,1) {$\neg A$};
\node (na1val) [color=DarkOrchid] at ($(na1) - (0,0.4)$) {$0$};
\node (b) at (3.5,1) {$B$};
\node (bval) [color=DarkOrchid] at ($(b) - (0,0.4)$) {$0$};
\node (nb1) at (4.2,1) {$\neg B$};
\node (nb1val) [color=DarkOrchid] at ($(nb1) - (0,0.4)$) {$0$};

\draw (or1.input 1) -- ++ (down: 0.25) -| (and1);
\draw (or1.input 2) -- ++ (down: 0.25) -| (and2);

\draw (and1.input 1) -- ++ (down: 0.25) -| (c);
\draw (and1.input 2) -- (or2);

\draw (and2.input 2) -- ++ (down: 0.25) -| (nc);
\draw (and2.input 1) -- (or3);

\draw (or2.input 2) -- (and3);

\draw (or3.input 1) -- ++ (down: 0.25) -| (and4);
\draw (or3.input 2) -- ++ (down: 0.25) -| (and5);

\draw (and3.input 1) -- ++ (down: 0.25) -| (or4);
\draw (and3.input 2) -- ++ (down: 0.25) -| (or5);

\draw (and4.input 1) edge (a);
\draw (and4.input 2) -- ++ (down:0.8) -| (nb);

\draw (and5.input 1) -- ++ (down:0.55) -| (na);
\draw (and5.input 2) -- ++ (down:0.25) -| (or5);

\draw (or4.input 1) -- ++ (down:0.25) -| (a1);
\draw (or4.input 2) edge (na1);

\draw (or5.input 1) edge (b);
\draw (or5.input 2) -- ++ (down:0.25) -| (nb1);
\end{tikzpicture}
}
\end{subfigure}
\caption{
For a given data point, the network (middle) outputs a distribution over classes $A, B$ and $C$, highlighted in blue, green and red, respectively.
The circuit encodes the constraint $(A \land B) \implies C$.
For each leaf node $l$, we plug in $\prob(l)$ and $1 - \prob(l)$ for positive and negative literals, respectively.
The computation proceeds bottom-up, taking products at AND gates and summations at OR gates.
The value accumulated at the root of the circuit (left) is the probability allocated by the network to the constraint.
The weights accumulated on edges from OR gates to their children are of special significance: OR nodes induce a partitioning of the distribution's support, and the weights correspond to the mass allocated by the network to each mutually-exclusive event.
Complemented with a second upward pass, where the entropy of an OR node is the entropy of the distribution over its children plus the expected entropy of its children, and the entropy of an AND node is the product of its children's entropies, we get the entropy of the distribution over the constraint's models -- the neuro-symbolic entropy regularization loss (right).
}
\label{fig:example}
\end{figure}
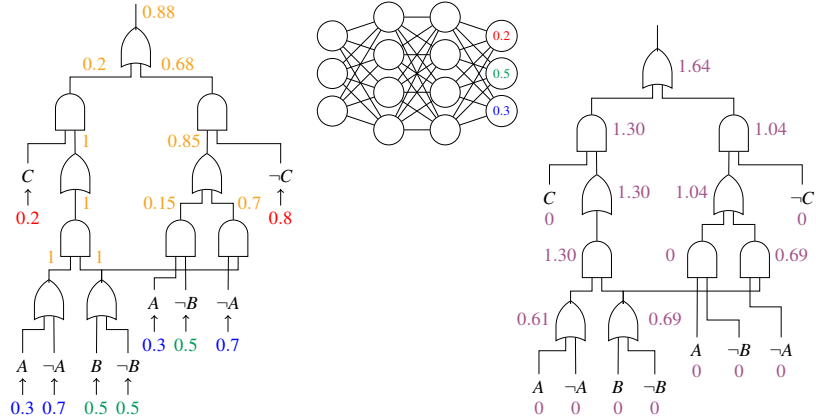
\subsection{An Illustrative example}\label{sec:example}
Consider Figure \ref{fig:example}.
Given a data point, the neural network defines a distribution over Boolean random variables $A, B$, and $C$, where $\prob(A) = \pv_0$ and $\prob(\lnot A) = 1 - \pv_0$, $\prob(B) = \pv_1$ and $\prob(\lnot B) = 1 - \pv_1$, etc.
The circuit encodes the constraint $(A \land B) \implies C$.
To compute the the probability of the constraint under the network's distribution, we feed the probabilities into the circuit, proceeding in a bottom-up fashion, taking products at AND gates and summations at OR gates, accumulating intermediate computations on the edges of the circuit.
The value accumulated at the root of the circuit is the probability mass allocated by the network to models of the formula, and corresponds to the probability of the constraint under the network's distribution -- this is exactly the semantic loss, up to a negative logarithm.
The weights accumulated on edges from OR gates to their children are of special significance: OR nodes induce a partitioning of the distribution's support, and the weights correspond to the mass allocated by the network to each mutually-exclusive event.
Complemented with another upward pass, where the entropy of every OR node is the entropy of the distribution over it's children plus the expected entropy of its children, and the entropy of every AND node is the product of its children's entropies, we calculate the entropy of the distribution over models of the constraint---precisely the neuro-symbolic entropy regularization. Therefore, performing two upward sweeps of the circuit, we can compute the neuro-symbolic entropy regularization and the semantic loss.

\subsection{Experimental Evaluation}\label{sec:experiments}

\subsubsection{Semi-Supervised: Entity-Relation Extraction}

\newcommand{\tablescale}{0.725}

\begin{table*}[!htb]
\caption{Experimental results for entity-relation extraction on ACE05 and SciERC. \#Labels indicates the number of labeled data points available to the network per relation. The remaining training set is stripped of labels and utilized in an unsupervised manner. We report the F1-score where a prediction is correct if the relation and its entities are correct.}
\centering
\scalebox{\tablescale}{%
\scriptsize
\begin{tabular}{llc|c|c|c|c|c|c}
\toprule
\# Labels  &    &3  &5  &10 &15 &25 &50 &75\\
\midrule
\multirow{6}{*}{\rotatebox[origin=c]{90}{ACE05}}
& Baseline
&4.92 $\pm$ 1.12          
&7.24 $\pm$ 1.75          
&13.66 $\pm$ 0.18 
&15.07 $\pm$ 1.79          
&21.65 $\pm$ 3.41          
&28.96 $\pm$ 0.98         
&33.02 $\pm$ 1.17 \\
& Self-training
&7.72 $\pm$ 1.21         
&12.83 $\pm$ 2.97            
&16.22 $\pm$ 3.08            
&17.55 $\pm$ 1.41            
&27.00 $\pm$ 3.66            
&32.90 $\pm$ 1.71           
&37.15 $\pm$ 1.42 \\
& Product t-norm
&8.89 $\pm$ 5.09       
&14.52 $\pm$ 2.13            
&19.22 $\pm$ 5.81            
&21.80 $\pm$ 7.67            
&30.15 $\pm$ 1.01           
&34.12 $\pm$ 2.75          
&37.35 $\pm$ 2.53 \\
\cmidrule{2-9}
& Semantic Loss
&12.00 $\pm$ 3.81
&14.92 $\pm$ 3.14 %
&22.23 $\pm$ 3.64
&27.35 $\pm$ 3.10
&30.78 $\pm$ 0.68
&36.76 $\pm$ 1.40
&38.49 $\pm$ 1.74\\
& + Full Entropy
&{\bf14.80} $\pm$ 3.70
&15.78 $\pm$ 1.90
&23.34 $\pm$ 4.07 %
&28.09 $\pm$ 1.46 %
&31.13 $\pm$ 2.26
&36.05 $\pm$ 1.00
&39.39 $\pm$ 1.21\\
& + NeSy Entropy
&14.72 $\pm$ 1.57
&{\bf18.38} $\pm$ 2.50
&{\bf26.41} $\pm$ 0.49
&{\bf31.17} $\pm$ 1.68
&{\bf35.85} $\pm$ 0.75
&{\bf37.62} $\pm$ 2.17
&{\bf41.28} $\pm$ 0.46\\
\midrule
\multirow{6}{*}{\rotatebox[origin=c]{90}{SciERC}}
& Baseline
&2.71 $\pm$ 1.10
&2.94 $\pm$ 1.00
&3.49 $\pm$ 1.80
&3.56 $\pm$ 1.10
&8.83 $\pm$ 1.00
&12.32 $\pm$ 3.00
&12.49 $\pm$ 2.60\\
&Self-training
&3.56 $\pm$ 1.40
&3.04 $\pm$ 0.90
&4.14 $\pm$ 2.60
&3.73 $\pm$ 1.10
&9.44 $\pm$ 3.80
&14.82 $\pm$ 1.20
&13.79 $\pm$ 3.90\\
&Product t-norm
&{\bf6.50} $\pm$ 2.00
&8.86 $\pm$ 1.20
&10.92 $\pm$ 1.60
&13.38 $\pm$ 0.70
&13.83 $\pm$ 2.90
&19.20 $\pm$ 1.70
&19.54 $\pm$ 1.70\\
\cmidrule{2-9}
&Semantic Loss
&6.47 $\pm$ 1.02
&{\bf9.31} $\pm$ 0.76
&11.50 $\pm$ 1.53
&12.97 $\pm$ 2.86
&14.07 $\pm$ 2.33
&20.47 $\pm$ 2.50
&23.72 $\pm$ 0.38
\\
&+ Full Entropy
&6.26 $\pm$ 1.21
&8.49 $\pm$ 0.85
&11.12 $\pm$ 1.22
&14.10 $\pm$ 2.79 
&17.25 $\pm$ 2.75
&{\bf22.42} $\pm$ 0.43
&24.37 $\pm$ 1.62\\
&+ NeSy Entropy
&6.19 $\pm$ 2.40
&8.11 $\pm$ 3.66
&{\bf13.17} $\pm$ 1.08
&{\bf15.47} $\pm$ 2.19
&{\bf17.45} $\pm$ 1.52
&22.14 $\pm$ 1.46
&{\bf25.11} $\pm$ 1.03\\
\bottomrule
\end{tabular}%
}
\label{table:results}
\end{table*}

Semantic loss and neuro-symbolic entropy were tested in a semi-supervised setting.
Here the model is presented with only a portion of the labeled
training set, with the rest used exclusively in an unsupervised manner
by the respective approaches.

We make use of the natural ontology of entity types and their relations 
present when dealing with relational data. This defines a set of 
relations and their permissible argument types. As is with all of our
constraints, we express the aforementioned ontology in the language 
of Boolean logic.

Our approach to recognizing the named entities and their pairwise relations
is most similar to \cite{Zhong2020}.
Contextual embeddings are first procured
for every token in the sentence. These are then fed into a named entity 
recognition module that outputs a vector of per-class probability for every 
entity. A classifier then classifies the concatenated contextual 
embeddings and entity predictions into a relation.

We employ two entity-relation extraction datasets, the Automatic Content
Extraction (ACE) 2005 \cite{walker2006} and SciERC datasets \cite{luan2018}.
ACE05 defines an ontology over $7$ entities and $18$ relations from mixed-genre
text, whereas SciERC defines $6$ entity types with $7$ possible relation between
them and includes annotations for scientific entities and there relations,
assimilated from $12$ AI conference/workshop proceedings.
We report the percentage of coherent predictions: data points for which the
predicted entity types, as well as the relations are correct.

We compare against five baselines. The first baseline is a purely supervised
model which makes no use of unlabeled data. The second is a classical
self-training approach based off of
\cite{chang2007}, and uses integer linear
programming to impute the unlabeled data's most likely labels subject to the
constraint, and consequently augment the (small) labeled set. The third 
baseline is a popular instantiation of a broad class of methods, fuzzy logics,
which replace logical operators with their fuzzy t-norms and logical implications
with simple inequalities. Lastly, we compare our proposed method, dubbed 
``NeSy Entropy'', to vanilla semantic loss
as presented in the earlier section
as well as another entropy-regularized baseline, dubbed ``Full Entropy'', which
minimizes the entropy of the entire predictive distribution, as opposed to just
the distribution over the constraint's models.

Our results are shown in Table \ref{table:results}. We observe that semantic loss outperforms
the baseline, self-training, and product t-norm across the board. We attribute
such a performance to the exactness of semantic loss, and its faithfulness to
the underlying constraint. We also observe that entropy-regularizing the
predictive model, in conjunction with training using semantic loss leads to
better predictive models, as compared with models trained solely using semantic
loss. Furthermore, it turns out that restricting entropy to the distribution
over the constraint's models, models that we know constitute the set of valid
predictions, compared to the model's entire predictive distribution, which
includes valid and invalid predictions, leads to a non-trivial
increase in the accuracy of predictions.

\newcommand{\ra}[1]{\renewcommand{\arraystretch}{#1}}

\begingroup
\begin{table}[t]
\ra{1.05}
\centering
\caption {Test results for grids, preference learning, and warcraft}
\label{tab:gridres}
\scalebox{\tablescale}{\small
\begin{tabular}{ @{} l l c c c @{}}
\toprule
& Test accuracy \%  & Coherent & Incoherent & Constraint \\
\midrule
\multirow{4}{15pt}{\rotatebox[origin=c]{90}{Grid}} & 5-layer MLP & \phantom{0}5.6 & {\bf 85.9} & \phantom{0}7.0 \\
\cmidrule{2-5}
& Semantic loss & 28.5 & 83.1 & 69.9 \\
& + Full Entropy & 29.0 & 83.8  & 75.2 \\ 
& + NeSy Entropy & {\bf 30.1} & 83.0 & {\bf 91.6} \\
\midrule
\multirow{4}{15pt}{\rotatebox[origin=c]{90}{Preference}} & 3-layer MLP & \phantom{0}1.0 & {\bf 75.8} & \phantom{0}2.7 \\
\cmidrule{2-5}
& Semantic loss & 15.0 & 72.4 & 69.8 \\
& + Full Entropy & 17.5 & 71.8 & 80.2 \\
& + NeSy Entropy & {\bf 18.2} & 71.5 & {\bf 96.0} \\
\midrule
\multirow{4}{15pt}{\rotatebox[origin=c]{90}{Warcraft}} & ResNet-18&  44.8 & 97.7  & 56.9\\
\cmidrule{2-5}
& Semantic loss& 50.9 &  97.7 & 67.4\\
& + Full Entropy& 51.5 &  97.6&  67.7\\
& + NeSy Entropy & {\bf55.0}& {\bf97.9}& {\bf69.8}\\
\bottomrule
\end{tabular}
}
\label{tab:grid}
\label{tab:pref}
\label{tab:sp}
\end{table}

\subsubsection{Fully-Supervised Learning}
Semantic loss and neuro-symbolic entropy were also tested in a fully supervised setting,
where our aim is to examine the effect of constraints enforced on the training set.
We note that this is a seemingly harder setting in the following sense: In a semi-
supervised setting we might make the argument that, despite its abundance, imposing
an auxiliary loss on unlabeled data provides the predictive model with an unfair
advantage as compared to the baseline. We concern ourselves with two tasks: predicting paths in a grid and preference learning.

\paragraph{Predicting Simple Paths}
For this task, our aim is to find the shortest path in a graph, or more
specifically a 4-by-4 grid, $G = (V, E)$ with uniform edge weights. Our input is
a binary vector of length $|V| + |E|$, with the first $|V|$ variables indicating
the source and destination, and the next $|E|$ variables encoding a subgraph $G'
\subseteq G$. Each label is a binary vector of length $|E|$ encoding the
shortest \emph{simple} path in $G'$, a requirement that we enforce through our
constraint. We follow the algorithm proposed by
\cite{nishino2017} to generate 
a constraint for each simple path in the grid, conjoined with indicators specifying
the corresponding source-destination pair. Our constraint is then the disjunction of all such conjunctions.

To generate the data, we begin by randomly removing one third of the edges in
the graph $G$, resulting in a subgraph, $G'$. Subsequently, we filter out
connected components in $G'$ with fewer than $5$ nodes to reduce degenerate
cases. We then sample a source and destination node uniformly at random. The
latter constitutes a single data point. We generate a dataset of $1600$
examples, with a $60/20/20$ train/validation/test split. %

\paragraph{Preference Learning}
We also consider the task of preference learning. Given the user's ranking of a subset of elements, we
wish to predict the user's preferences over the remaining elements of the set.
We encode an ordering over $n$ items as a binary matrix ${X_{ij}}$, where for
each $i, j \in {1, \ldots, n}$, $X_{ij}$ denotes that item $i$ is at position
$j$. Our constraint $\alpha$ requires that the network's output be a valid 
total ordering.
We use preference ranking data over $10$ types of sushi for $5,000$ individuals,
taken from PREFLIB \cite{MaWa13a}, split 60/20/20. Our inputs consist of the user's 
preference over $6$ sushi types, with the model tasked to predict the
user's preference, a \emph{strict} total order, over the remaining $4$. %

Tables \ref{tab:grid} compares the baseline
to the same MLP augmented with semantic loss, semantic
loss with entropy regularization over the entire predictive distribution, dubbed
``Full Entropy'' and entropy regularization over the distribution over the
constraint's models, dubbed ``NeSy Entropy".

Similar to \cite{Xu18}, we observe that the semantic loss has a marginal effect
on incoherent accuracy, but significantly improves the network’s  ability to output
coherent predictions. We also observe that, similar to semi-supervised settings, 
entropy-regularization leads to more coherent predictions using
both ``Full Entropy'' and ``NeSy Entropy", with ``NeSy Entropy" leading to the
best performing predictive models. Remarkably, we also observe that ``NeSy Entropy''
leads to predictive models whose predictions almost always satisfy the constraint,
captured by ``Constraint''.

\paragraph{Warcraft Shortest Path}
Lastly, we consider a more real-world variant of the task of predicting simple paths.
Following \cite{Pogancic2020}, our training set consists of $10,000$
terrain maps curated using Warcraft II tileset.
Each map encodes an underlying grid of dimension $12 \times 12$, where each vertex is assigned a cost depending on the type of terrain it represents (e.g. earth has lower cost than water).
The shortest (minimum cost) path between the top left and bottom right vertices is encoded as
an indicator matrix, and serves as label. 
Figure~\ref{fig:sp-results} shows an example input presented to the network, the groundtruth, and the input with the annotated shortest path.

\setlength{\fboxsep}{0pt}
\setlength{\picHeight}{0.20\linewidth}
\begin{figure}[t]Lastly
    \centering
        \scalebox{0.9}{
		\parbox[b][\picHeight][c]{1em}{\rotatebox{90}{Input}}
		\includegraphics[height=\picHeight]{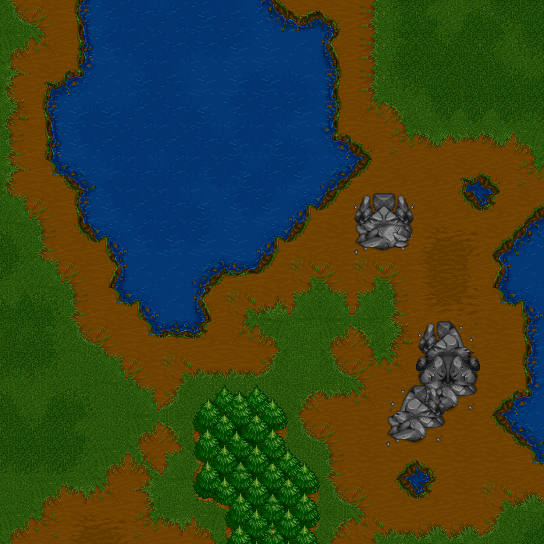}
		\vbox to\picHeight{\vfil\hbox{~~~~\LARGE$\to$}\vfil}
		\parbox[b][\picHeight][c]{\picHeight}{
		\centering
		$$
			\begin{pmatrix}
				1&0&\cdots&0\\
				1&0&\cdots&0\\
				\vdots & \vdots & \ddots &\vdots\\
				0&0&\cdots&1
			\end{pmatrix}
		$$} ~~~~~~
		\includegraphics[height=\picHeight]{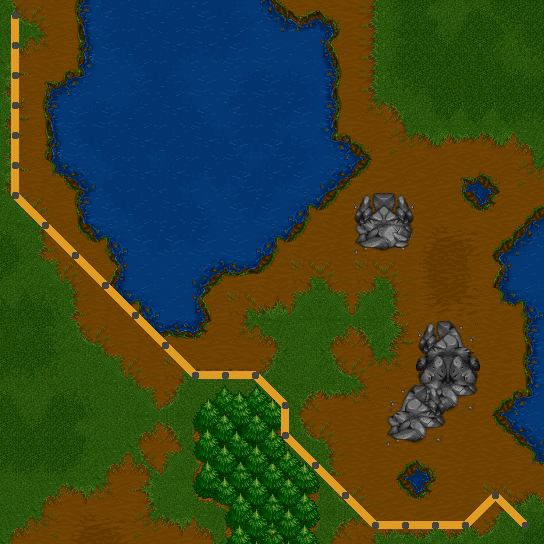}
                }
		\caption{Warcraft dataset. Each
		input (left) is a $12 \times 12$ grid corresponding to a Warcraft II
		terrain map, the output is a matrix (middle) indicating the shortest
		path from top left to bottom right (right).
		\label{fig:sp-results:dataset}
		\label{fig:sp-results:withpath}
		}
    \label{fig:sp-results}
\end{figure}

Presented with an image of a terrain map, a convolutional neural network -- following \cite{Pogancic2020}, we use ResNet18 \cite{He2016} -- outputs a $12 \times 12$ binary matrix indicating the vertices that constitute the minimum cost path.
We report three metrics: ``Coherent'' denotes the percentage of optimal-cost predictions, ``Incoherent'' denotes the percentage of individual vertices matching the groundtruth, and ``Constraint'' indicates the percentage of predictions that constitute valid paths. Our results are shown in Table~\ref{tab:sp}.

In line with our previous experiments, we observe that incorporating constraints into learning improves the ``Coherent'' metric from $44.8\%$ to $50.9\%$, and of the  ``Coherent'' metric from $56.9\%$ to $67.4\%$.
Augmenting semantic loss with the entropy over the network's predictive distribution, ``Full Entropy'', we attain a modest improvement from $50.9\%$ to $51.5\%$ and $67.4\%$ to $67.7\%$ for the ``Coherent'' and ``Constraint'' metrics respectively. 
Restricting the entropy minimization to models of the constraint, ``NeSy Entropy'', we observe that we attain a large improvement to $55.0\%$ and $69.8\%$ for the ``Coherent'' and ``Constraint'' metrics resp.

\section{Generating Structures with Constrained Adversarial Networks}
\label{sec:can}

In this section, we discuss how the Semantic Loss can be implemented to design deep generative models able to output realistic structured objects, focusing on the case of Generative Adversarial Networks (GANs)~\cite{goodfellow2014generative}.
GANs are composed of two neural networks:
a discriminator $\dis$ trained to recognize ``real'' objects
$\x$ sampled from the data distribution $\pdata$, and
a generator $\gen$ that maps random latent vectors $\z$
to objects $g(\x)$ that fool the discriminator.
Learning equates to solving the minimax game $\min_{\gen} \max_{\dis} \;
f_\text{GAN}(\gen, \dis)$ with value function:
\begin{equation}
    f_\text{GAN}(\gen, \dis) := \E_{\x \sim \pdata}[\log \pdis(\x)] + \E_{\x \sim \pgen}[\log (1 - \pdis(\x))]
    \label{eq:gan}
\end{equation}
Here $\pgen(\x)$ and $\pdis(\x) := \pdis(\text{real}\,|\,\x)$ are the
distributions induced by the generator and discriminator, respectively.
New objects $\x$ can be sampled by mapping random vectors $\z$
using the generator, i.e., $\x = \gen(\z)$.
It is well known that, under idealized assumptions, the learned generator matches the data
distribution:

\begin{thm}[\cite{goodfellow2014generative}]
    \label{thm:gan}

    If \gen and \dis are non-parametric and the leftmost
    expectation in Eq.~\ref{eq:gan} is approximated arbitrarily well by the
    data, the global equilibrium $(\gen^*, \dis^*)$ of
    Eq.~\ref{eq:gan} satisfies $\prob_{\dis^*} \equiv \frac{1}{2}$ and
    $\prob_{\gen^*} \equiv \pdata$.

\end{thm}

In practice, training GANs is notoriously
hard~\cite{salimans2016improved,mescheder2018training}.  The most
common failure mode is mode collapse, in which the generated objects
are clustered in a tiny region of the object
space.
Remedies include using alternative objective
functions~\cite{goodfellow2014generative},
divergences~\cite{nowozin2016f,arjovsky2017wasserstein} and
regularizers~\cite{miyato2018spectral}.

In structured tasks, which are our main focus, the objects of interest are usually discrete.  In
the following, we focus on stochastic generators that output
a \emph{categorical distribution} $\vtheta(\z)$ over $\Xs$ and
objects are sampled from the latter.
In this case,
$
    \pgen(\x) = \int \pgen(\x | \z) p(\z) d\z = \int \vtheta(\z) p(\z) d\z = \E_{\z}[\vtheta(\z)]
$.

Alas, standard GANs struggle to output valid structures, for two main reasons.
First, the number of examples necessary to capture any non-trivial constraint $\alpha$ can be intractably large.\footnote{The VC dimension of unrestricted discrete formulas is exponential in the number of variables~\cite{vapnik2015uniform}.}  This makes it highly non-trivial to learn the underlying constraints -- such as the rules of chemical validity or, worse still, graph reachability -- from even moderately large data sets.  Second, in many cases of interest the examples are noisy and \textit{do} violate $\alpha$, in which case the data lures GANs into learning \emph{not} to satisfy the constraint.  This point is made more explicit in the next theorem:

\begin{cor}
    \label{thm:ideal_bad_gan}
    Under the assumptions of Theorem~\ref{thm:gan}, given a target distribution \pdata, a constraint $\alpha$ consistent with it, and a dataset of examples $\x$ sampled i.i.d. from a corrupted distribution $\pdata' \ne \pdata$ inconsistent with $\alpha$, GANs associate non-zero mass to infeasible objects.
\end{cor}

This result follows immediately from Theorem~\ref{thm:gan}, as the optimal generator satisfies $\pgen \equiv \pdata'$, which is inconsistent with $\alpha$.  Since Theorem~\ref{thm:gan} captures the \emph{intent} of GAN training, this corollary shows that GANs are \emph{by design} incapable of handling invalid examples.

\subsection{Constrained Adversarial Networks}

Constrained Adversarial Networks (CANs) are a deep generative model that outputs structures $\x$ consistent with validity constraints and an unobserved distribution \pdata.
We assume to be given:
i)~a feature map $\phi$ that extracts $b$ binary features from $\x$, and
ii)~a single validity constraint $\alpha$ encoded as a Boolean formula on $\phi(\x)$.
If $\x$ is binary, $\phi$ can be taken to be the identity; later we will discuss some alternatives.
Any discrete structured space can be encoded this way.

CANs avoid the brittleness of regular GANs by taking both the target structural constraint $\alpha$ as inputs and adapting the value function so that the generator maximizes the probability of generating valid structures.  In order to derive CANs it is convenient to start from the following alternative GAN value function~\cite{goodfellow2014generative}:
$
    f_\text{ALT}(\gen, \dis) := \E_{\x \sim \pdata}[\log \pdis(\x)] - \E_{\x \sim \pgen}[\log \pdis(\x)]
    \label{eq:altgan}
$.
Now, let $(\gen, \dis)$ be a GAN and $\val(\x) = \Ind{\phi(\x) \models \alpha}$ be a fixed discriminator that distinguishes between valid and invalid structures.  Ideally, we wish the generator to \emph{never} output invalid structures.  This can be achieved by using an aggregate discriminator $\valdis(\x)$ that only accepts configurations that are both valid and high-quality w.r.t. $\dis$.  Let $A$ be the indicator that $\valdis$ classifies $\x$ as real, and similarly for $D$ and $V$.  By definition:
\begin{equation}
    \prob_\valdis(\x)
    = \prob(A \,|\, \x)
    = \prob(D \,|\, V, \x) \prob(V \,|\, \x)
    = \pdis(\x) \Ind{\phi(\x) \models \alpha}
    \label{eq:valdis}
\end{equation}
Plugging the aggregate discriminator into the alternative value function gives:
\begin{align}
    & \argmax_\valdis \; f_\text{ALT}(\gen, \valdis)
    \\
    & = \argmax_\dis \; \E_\pdata[\log \pdis(\x) + \log \Ind{\phi(\x) \models \alpha}] - \E_\pgen[\log \pdis(\x) + \log \Ind{\phi(\x) \models \alpha}]
    \\
    & = \argmax_\dis \; \E_\pdata[\log \pdis(\x)] - \E_\pgen[\log \pdis(\x)] - \E_\pgen[\log \Ind{\phi(\x) \models \alpha}]
    \\
    & = \argmax_\dis \; f_\text{ALT}(\gen, \dis) - \E_\pgen[\log \Ind{\phi(\x) \models \alpha}]
    \label{eq:altcan}
\end{align}
The second step holds because $\E_\pdata[\log \Ind{\phi(\x) \models \alpha}]$ does not depend on \dis.  If \gen allocates non-zero mass to \emph{any} measurable subset of invalid structures, the second term becomes $+\infty$. This is consistent with our goal but problematic for learning.  A better alternative is to optimize the lower bound:
\begin{equation}
    \SL{\alpha}{\gen}
    :=
    -\log \pgen(\alpha)
    =
    - \log \E_\pgen[\Ind{\phi(\x) \models \alpha}]
    \le
    - \E_\pgen[\log \Ind{\phi(\x) \models \alpha}]
    \label{eq:lowerbound}
\end{equation}
This term is exactly the semantic loss.  In this context, the semantic loss represents a relaxation of the original, in the sense that it only evaluates to $+\infty$ if \pgen allocates \emph{all} the mass to infeasible configurations.  This immediately leads to the CAN value function:
\begin{equation}
    \textstyle
    f_\text{CAN}(\gen, \dis) := f_\text{ALT}(\gen, \dis) + \lambda \SL{\alpha}{\gen}
    \label{eq:can}
\end{equation}
where $\lambda > 0$ is a hyper-parameter controlling the importance of the constraint.  This formulation is related to integral probability metric-based GANs, cf.~\cite{li2017mmd}.  The SL can be viewed as the negative log-likelihood of $\alpha$, and hence it rewards the generator proportionally to the mass it allocates to valid structures.  The expectation in Eq.~\ref{eq:lowerbound} can be rewritten as:
\begin{equation}
    \E_{\x \sim \pgen}[\Ind{\phi(\x) \models \alpha}]
    = \sum_{\x : \phi(\x) \models \alpha} \pgen(\x)
    = \E_{\z}\left[ \sum_{\x : \phi(\x) \models \alpha} \: \prod_{i \,:\, x_i = 1} \theta_i(\z) \!\! \prod_{i \,:\, x_i = 0} (1 - \theta_i(\z)) \right]
    \label{eq:wmc}
\end{equation}
Hence, the SL is the negative logarithm of a polynomial in $\vtheta$ and it is fully differentiable.\footnote{As long as $P_\gen(\x) > 0$, which is always the case in practice.}  In practice, below we apply the semantic loss term directly to $f_\text{GAN}$, i.e., $f_\text{CAN}(\gen,\dis) := f_\text{GAN}(\gen,\dis) + \lambda \SL{\alpha}{\gen}$.

If the SL is given large enough weight $\lambda$ then it gets closer to the ideal ``hard'' discriminator, and therefore more strongly encourages the CAN to generate valid structures.  Under the preconditions of Theorem~\ref{thm:gan}, it is clear that for $\lambda \to \infty$ CANs generate valid structures only:

\begin{prop}
    \label{thm:ideal_can}
    Under the assumptions of Corollary~\ref{thm:ideal_bad_gan}, CANs associate
    zero mass to infeasible objects, irrespective of the discrepancy between
    \pdata\ and $\pdata'$.
\end{prop}

Indeed, any global equilibrium $(\gen^*, \dis^*)$ of $\min_\gen \max_\dis f_\text{CAN}(\gen, \dis)$ minimizes the second term: the minimum is attained by $\log \prob_{\gen^*}(\alpha) = 0$, which entails $\prob_{\gen^*}(\lnot \alpha) = 0$.
Of course, as with standard GANs, the prerequisites are often violated in practice.  Regardless, Proposition~\ref{thm:ideal_can} works as a sanity check, and shows that, in contrast to GANs, CANs are appropriate for structured generative tasks.

\subsection{Level Generation}

In order to test the benefits of the semantic loss, we applied CANs applied to Super Mario Bros. (SMB) level generation. SMB levels
are tile-based and can be encoded by a matrix of categorical values
(such as \textit{wall}, \textit{ground}, \textit{coin}, etc.).
Pipes are made out of four different types of tiles. They can have a
variable height but the general structure is always the same: two
tiles (\textit{top-left} and \textit{top-right}) on top and one or
more pairs of body tiles (\textit{body-left} and \textit{body-right})
below (see the \texttt{CAN - pipes} in picture in
Fig.~\ref{fig:smb-ex} for examples of valid pipes).
The structural constraint on pipes consists of multiple implications
like ``if this is a \textit{top-left} tile, then the tile below must
be a \textit{body-left} one'' conjoined together. Noticeably, these
structural constraints can be locally defined on a $2 \times 2$
portion of the level. The loss was thus computed globally by applying
the same constraint over a sliding window, ultimately requiring a
significantly more compact circuit.
Two major problems were observed when using a large $\lambda$:
i)~\textit{vanishing pipes}: this occurs because the generator can satisfy the constraint by simply generating layers without pipes;
ii)~\emph{mode collapse}: the generator may learn to place pipes always in the same positions.
In this setting, both issues were addressed by introducing the SL
after an initial bootstrap phase in which the generator learns to
generate sensible objects, and by linearly increasing $\lambda$.
Table~\ref{tab:pipes-res} reports the experimental results of the
comparison between a baseline GAN and CAN trained on all levels
containing pipes from the \emph{video game level corpus}
(VGLC)~\cite{summerville2016vglc}. CANs managed to almost double the
validity of the generated levels (see the two left pictures in
Fig.~\ref{fig:smb-ex} for some prototypical examples) while retaining
about 82\% of the pipe tiles and without any significant loss in terms
of diversity (as measured by the L1 norm on the difference between
each pair of levels in the generated batch) or
inference time.

\begin{figure*}[tb]
\begin{center}
\begin{tabular}{cccc}
  {\small \texttt{GAN - pipes}} &
  {\small \texttt{CAN - pipes}} &
  {\small  \texttt{GAN - playability}} &
  {\small \texttt{CAN - playability}} \\
  \includegraphics[width=0.22\linewidth]{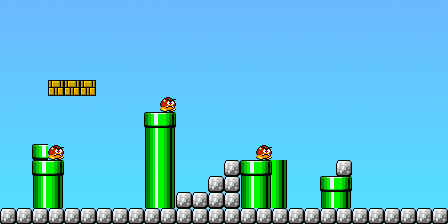} &
  \includegraphics[width=0.22\linewidth]{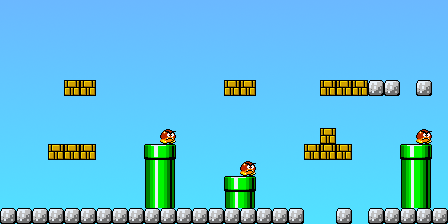} &
  \includegraphics[width=0.22\linewidth]{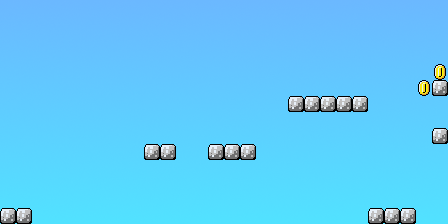} &
  \includegraphics[width=0.22\linewidth]{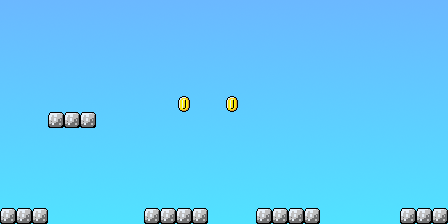} \\
\end{tabular}
\end{center}
\caption{Examples of SMB levels generated by GAN and CAN. Left: generating levels containing pipes; right: generating reachable levels. For each of the two settings we report prototypical examples of levels generated by GAN (first and third picture) and CAN (second and fourth picture). Notice how all pipes generated by CAN are valid, contrarily to what happens for GAN, and that the GAN generates a level that is not playable (because of the big jump at the start of the map).}

\label{fig:smb-ex}
\end{figure*}

\begin{table}[tb]
  \centering

  \caption{\label{tab:pipes-res} Comparison between GAN and CAN on SMB level generation with pipes.
Results report validity, average number of pipe tiles per level, L1 norm on the difference between each pair of levels in the generated batch and training time. Inference is real-time ($< 40$ ms) for both architectures.}
	\tiny
	\begin{tabular}{cccccc}
	\toprule
	\textbf{Model} & \textbf{\# Maps} & \textbf{Validity (\%)} &
	    \textbf{Average pipe-tiles / level} & \textbf{L1 Norm} & \textbf{Training time} \\
    \midrule
	GAN & 7 & 47.6 $\pm$ 8.3 & 7.8   & 0.0115    & 1h 12m \\
	CAN & 7 & 83.2 $\pm$ 4.8 & 6.4   & 0.0110    & 2h 2m \\
	\bottomrule
\end{tabular}
\end{table}

\subsection{Conditional CANs}

So far we described how to use the SL for enforcing structural constraints on the generator's output. Since the SL can be applied to any distribution over binary variables, it can also be used to enforce conditional constraints that can be turned on and off at inference time. Specifically, we notice that the constraint can involve also latent variables, and we show how this can be leveraged for different purposes.
Similarly to InfoGANs~\cite{chen2016infogan}, the generator's input is augmented with an additional binary vector $\vc$. Instead of maximizing (an approximation of) the mutual information between $\vc$ and the generator's output, the SL is used to logically bind the input codes to semantic features or constraint of interest.
Let $\alpha_1, \ldots, \alpha_k$ be $k$ constraints of interest.  In order to make them switchable, we extend the latent vector $\z$ with $k$ fresh variables $\vc = (c_1, \ldots, c_k) \in \{0,1\}^k$ and train the CAN using the constraint:
\[
    \textstyle
    \alpha = \bigwedge_{i=1}^k (c_i \leftrightarrow \alpha_i)
    \label{eq:onoff}
\]
where the prior $P(\vc)$ used during training is estimated from data.
Using a conditional SL term during training results in a model that can be conditioned to generate object with desired, arbitrarily complex properties $\alpha_i$ at inference time.

\subsection{Molecule Generation}

An experiment on generating molecules in graphical form investigated
the use of conditional CANs in conjunction with other forms of
regularization.  Specifically, MolGAN's adversarial training
and \emph{reinforcement learning objective}~\cite{de2018molgan} was
combined with a conditional SL term on the task of generating
molecules that maximize certain desirable chemical properties.
The structured objects are undirected graphs of bounded maximum size,
represented by discrete tensors that encode the atom/node type
(padding atom (no atom), Carbon, Nitogren, Oxygen, Fluorine) and the
bound/edge type (padding bond (no bond), single, double, triple and
aromatic bond).
During training, the RL objective implicitly rewarded validity and the
maximization of the three chemical properties at once: \textbf{QED}
(druglikeness), \textbf{SA} (synthesizability) and \textbf{logP}
(solubility).
The architecture was augmented with a conditional SL term, making
use of $4$ latent dimensions to control the presence of one of the $4$
types of atoms considered in the experiment.
Conditioning the generation of molecules with specific atoms at
training time mitigated the drop in uniqueness caused by the reward
network during the training. This allowed the model to be trained for
more epochs and resulted in more diverse and higher quality molecules,
as reported in Table~\ref{tab:mol}.

\begin{table}[tb]
  \centering
      \caption{\label{tab:mol} Results of using the semantic loss on the MolGAN
  architecture. The diversity score is obtained by comparing
  sub-structures of generated samples against a random subset of the
  dataset. A lower score indicates a higher amount of repetitions
  between the generated samples and the dataset.  The first row refers
  to the results reported in the MolGAN paper.
}
    \begin{footnotesize}
    \begin{tabular}{cccccccc}
    \toprule
        \textbf{Reward for} & \textbf{SL} & \textbf{validity} & \textbf{uniqueness} & \textbf{diversity} & \textbf{QED} & \textbf{SA}   & \textbf{logP} \\ \midrule
        \multirow{2}*{QED + SA + logP}
                    & False               & 97.4              & 2.4                 & 91.0               & 47.0         & 84.0          & 65.0          \\ \cline{2-8}
                    & True                & 96.6      & 2.5         & 98.8       & 51.8 & 90.7  & 73.6  \\ \bottomrule
    \end{tabular}
    \end{footnotesize}
    \vspace{0.8em}
\end{table}

\subsection{The Embedding Function $\phi$ and Large-scale Level Generation}

The embedding function $\phi(\x)$ extracts Boolean variables to which the SL is then applied.  In many cases, as in the experiments with pipes and molecules, $\phi$ is simply the identity map.  However, when fed a particularly complex constraint $\alpha$, KC may output too large a circuit.  In this case, $\phi$ can be used to map $\x$ to an application-specific embedding space where $\alpha$ (and hence the SL polynomial) is expressible in compact form.

Another experiment on SMB level generation showed how CANs can be
successfully applied in settings where constraints are too complex to
be directly encoded onto the generator output. In order for a level to
be playable, there must exist a traversable path\footnote{According to
the game's physics.} from the left-most to the right-most column of
the level. We refer to this property as {\em reachability}.
In this context, the prominent approach consists in training a
generative model without semantic constraints, using an evolutionary
search on the latent space when sampling objects with desired
properties at inference time. An advantage of this approach is that
the fitness function doesn't have to be differentiable. For instance,
reachability can be computed by letting a black-box A* agent play the
level. This key idea was applied to SMB level
generation~\cite{volz2018evolving} using the Covariance Matrix
Adaptation Evolution Strategy (CMA-ES).
Having the SL to steer the generation towards playable levels
is not trivial, since it requires a differentiable definition of
reachability.
Since directly encoding the constraint in propositional logic is
intractable, the reachability constraint was defined on the output of
an embedding function $\phi$, modelled as a feedforward NN, that
approximates tile reachability. The function was pre-trained to
predict whether each tile is reachable from the left-most column using
traces obtained from the A* agent.
The two approaches were compared, using the exact same GAN
architecture, training setting and A* agent. In particular, training
data was obtained by slicing a single SMB level from the VGLC.
Results showed that CANs achieves better (\textit{mario-1-3}) or
comparable (\textit{mario-3-3}) validity with respect to the baseline
GAN + CMA-ES at a fraction of the inference time. At the cost of
pretraining the reachability function, CANs avoid the execution of the
A* agent during the generation and sample high quality objects in
milliseconds (as compared to minutes), thus enabling applications to
create new levels at run time. Notably, no significant quality
degradation was observed with respect to the baseline GAN (which on
the other hand fails most of the time to generate reachable levels).

\begin{table}[tb]
  \centering
  \caption{\label{tab:smb-res} Results on the generation of {\em playable} SMB level. Levels
  \textit{mario-1-3} ($123$ training samples) and \textit{mario-3-3}
  ($122$ training samples) were chosen due to their
  high solving complexity. Results compare a baseline GAN, a GAN
  combined with CMA-ES and a CAN. Validity is defined as the ability
  of the A* agent to complete the level. Note that inference time for
  GAN and CAN is measured in milliseconds while time for GAN + CMA-ES
  is in minutes.}
	\tiny
	\begin{tabular}{cccccc}
	\toprule
	\textbf{Network type} & \textbf{Level} & \textbf{Tested samples} & \textbf{Validity} & \textbf{Training time} & \textbf{Inference time per sample} \\
	\midrule
	GAN &			mario-1-3	&	1000 &	9.80\% &		1 h 15 min &	$\sim$ 40 ms \\
	GAN + CMA-ES &	mario-1-3	&	1000 &	65.90\% &		1 h 15 min &	$\sim$ 22 min \\
	CAN &			mario-1-3	&	1000 &	71.60\% &	 	1 h 34 min & 	$\sim$ 40 ms \\
	\midrule
	GAN &			mario-3-3	&	1000 &	13.00\% &		1 h 11 min &	$\sim$ 40 ms \\
	GAN + CMA-ES &	mario-3-3	&	1000 &	64.20\% &		1 h 11 min & 	$\sim$ 22 min \\
	CAN &			mario-3-3   &	1000 &	62.30\% &		1 h 27 min & 	$\sim$ 40 ms \\
	\bottomrule
\end{tabular}

\end{table}

\section{Related Work}

In an acknowledgment to the need for both symbolic as well as sub-symbolic reasoning, there has been a plethora of recent works studying how to best combine neural networks and logical reasoning, dubbed \emph{neuro-symbolic reasoning}. The focus of such approaches is typically making probabilistic reasoning tractable through first-order approximations, and differentiable, through reducing logical formulas into arithmetic objectives, replacing logical operators with their fuzzy t-norms, and implications with inequalities~\cite{kimmig2012short,rocktaschel2015,fischer19a}. \cite{diligenti2017} and \cite{donadello2017} use first-order logic to specify constraints on outputs of a neural network. They employ fuzzy logic to reduce logical formulas into differential, arithmetic objectives denoting the extent to which neural network outputs violate the constraints, thereby supporting end-to-end learning under constraints. However, apart from lacking
a formal derivation in terms of expected probability of satisfying
constraints, the issue is that fuzzy logic is not semantically sound,
meaning that equivalent encodings of the same constraint may give
different loss functions~\cite{giannini2018convex}. The semantic loss instead preserves the semantics of the symbolic knowledge regardless of how the latter is encoded, avoiding these issues.
Recently, \cite{Ahmed22pylon} proposed using sampling to obtain a Monte Carlo estimate of the probability of the constraint being satisfied. This offers the convenience of specifying constraints as PyTorch functions, as well as accommodating non-differentiable elements in the training process.
However, estimating the probability using sampling can become infeasible in exponentially-sized output spaces where the valid outputs represent only a sliver of the distribution’s support, or the neural network’s outputs largely satisfy the constraint.
Semantic strengthening \cite{AhmedAISTATS23} approaches the problem by first assuming the constraint decomposes conditioned on the learned features, then iteratively strengthening the approximation, restoring the dependence between the constraints most responsible for degrading the quality of the approximation. This corresponds to computing the mutual information between pairs of constraints conditioned on the learned features, and may be construed as a measure of gradient alignment.

Another class of neuro-symbolic approaches have their roots in logic programming. DeepProbLog~\cite{manhaeve2018} extends ProbLog, a probabilistic logic programming language, with the capacity to process neural predicates, whereby the network's outputs are construed as the probabilities of the corresponding predicates. This simple idea retains all essential components of ProbLog: the semantics, inference mechanism, and the implementation. In a similar vein, \cite{dai2018} combine domain knowledge specified as purely logical Prolog rules with the output of neural networks, dealing with the network's uncertainty through revising the hypothesis by iteratively replacing the output of the neural network with anonymous variables until a consistent hypothesis can be formed. \cite{bosnjak2017programming} present a framework combining prior procedural knowledge, as a Forth program, with neural functions learned through data. The resulting neural programs are consistent with specified prior knowledge and optimized with respect to data.

There has recently been a plethora of approaches ensuring consistency by embedding the constraints as predictive layers, including semantic probabilistic layers (SPLs) \cite{Ahmed2022}, MultiplexNet~\cite{Hoernle2021MultiplexNetTF} and HMCCN~\cite{giunchiglia2020coherent}.
Much like semantic loss \cite{Xu18}, SPLs maintain sound probabilistic semantics, support general constraints expressed in propositional logic, but in addition, impose hard constraints on the network's output.
SIMPLE \cite{AhmedICLR2023} uses an SPL for the distribution over susbets of size $k$, which is shown to be a tractable distribution, for which it derives a gradient estimator as a function of the exact conditional marginals.
MultiplexNet is able to encode only constraints in disjunctive normal form, which is problematic for generality and efficiency as neuro-symbolic tasks often involve an intractably large number of clauses.
HMCCN encodes label dependencies as fuzzy relaxation and is the current state-of-the-art model for hierarchical mutli-label classification~\cite{giunchiglia2020coherent}, but, similar to its recent extension~\cite{giunchiglia2021multi}, is restricted to a certain family of constraints. %

The objective of neuro-symbolic entropy to increase the confidence of predictions on unlabeled data is related to information-theoretic approaches to semi-supervised learning~\cite{grandvalet2005, erkan2010}, and approaches that increase robustness to output perturbation ~\cite{Miyato2016}. A key difference between neuro-symbolic entropy and these information-theoretic losses is the former takes the structure of the  output space into account, while the latter does not.

\section{Conclusion}

Neural networks have achieved breakthroughs across a wide range of domains.
Even in the presence in an abundance of data, however, they still struggle 
to learn the intricate dependencies between labels in structured output spaces.
Semantic loss offers a remedy by penalizing the network for any probability mass
allocated to invalid outputs, ensuring the distribution only allows for predictions
that are consistent with the constraint.
Neuro-symbolic entropy goes one step further, requiring that the network learn a
minimum-entropy distribution consistent with the constraint.
The benefits of semantic loss are not restricted to structured-output prediction, however.
Constrained adversarial networks extend semantic loss to generative learning, maximizing the
probability that the generated outputs satisfy the constraint.

\section*{Acknowledgements}

The research of ST and AP was partially supported by TAILOR, a project funded by EU Horizon 2020 research and innovation programme under GA No 952215.
This work was funded in part by the DARPA Perceptually-enabled Task Guidance (PTG) Program under contract number HR00112220005, NSF grants \#IIS-1943641, \#IIS-1956441, \#CCF-1837129, Samsung, CISCO, a Sloan Fellowship, and a UCLA Samueli Fellowship. The research of PM received funding from the Flemish Government under the “Onderzoeksprogramma Artificiële Intelligentie (AI) Vlaanderen” programme and the Research Foundation - Flanders under the Data- driven logistics project (FWO-S007318N).

\bibliographystyle{tfnlm}
\bibliography{chapter-sl/reference}

\end{document}